\documentclass[conference]{IEEEtran}
\IEEEoverridecommandlockouts
\usepackage{cite}
\usepackage{amsmath,amssymb,amsfonts}
\usepackage{algorithmic}
\usepackage{graphicx}
\usepackage{textcomp}
\usepackage{xcolor}
\usepackage{authblk}

\usepackage{graphicx}
\usepackage{subcaption}

\def\BibTeX{{\rm B\kern-.05em{\sc i\kern-.025em b}\kern-.08em
    T\kern-.1667em\lower.7ex\hbox{E}\kern-.125emX}}
\begin{document}

\title{Spectral Pyramid Graph Attention Network for Hyperspectral Image Classification}

\author[1]{Tinghuai Wang}
\author[1]{Guangming Wang}
\author[1]{Kuan Eeik Tan}
\author[2]{Donghui Tan}
\affil[1]{Huawei Helsinki Research Center, Finland \authorcr Email: {\tt tinghuai.wang@huawei.com}\vspace{1.5ex}}
\affil[2]{Huawei Technologies, China}% \authorcr Email: {\tt uid3@jpl.nasa.gov} \vspace{-2ex}} 

%\author{\IEEEauthorblockN{Tinghuai Wang,
%Guangming Wang, Kuan Eeik Tan, Aimeng Wang and
%Donghui Tan}
%\IEEEauthorblockA{Department of Whatever,
%Whichever University\\
%Wherever\\
%Email: \{a1, a2, a3\}author.one@add.on.net}}

\maketitle

\begin{abstract}
Convolutional neural networks (CNN) have made significant advances in hyperspectral image (HSI) classification. However, standard convolutional kernel neglects 
the intrinsic connections between data points, resulting in poor region delineation and small spurious predictions. Furthermore, HSIs have a unique continuous data 
distribution along the high dimensional spectrum domain - much remains to be addressed in characterizing the spectral contexts 
considering the prohibitively high dimensionality and improving reasoning capability in light of the limited amount of labelled data. This paper presents a 
novel architecture which explicitly addresses these two issues. Specifically, we design an architecture to encode the multiple spectral contextual information
in the form of spectral pyramid of multiple embedding spaces. In each spectral embedding space, we propose graph attention mechanism to explicitly 
perform interpretable reasoning in the spatial domain based on the connection in spectral feature space. Experiments on three HSI datasets demonstrate that
the proposed architecture can significantly improve the classification accuracy compared with the existing methods.

%Existing CNN architectures are typically utilizing 
%standard convolutional kernels with fixed weights given their position within the convolution window. This limitation of the standard convolutional kernel neglects 
%the structural connections between data points, resulting in poor region delineation and small spurious predictions.

%Existing CNN architectures are typically designed and optimized for classifying RGB images, which excel in capturing the 2D spatial contexts while learning a rich feature representation. However, HSIs 
%normally have a unique nature of data distribution along the high dimentional spectrum domain - much remains to be addressed in capturing the  spectral contexts considering the prohibitively high dimensionality 
%and improving reasoning capability in light of the limited amount of labelled data. 

\end{abstract}


\begin{figure*}[htbp]
\centerline{\includegraphics{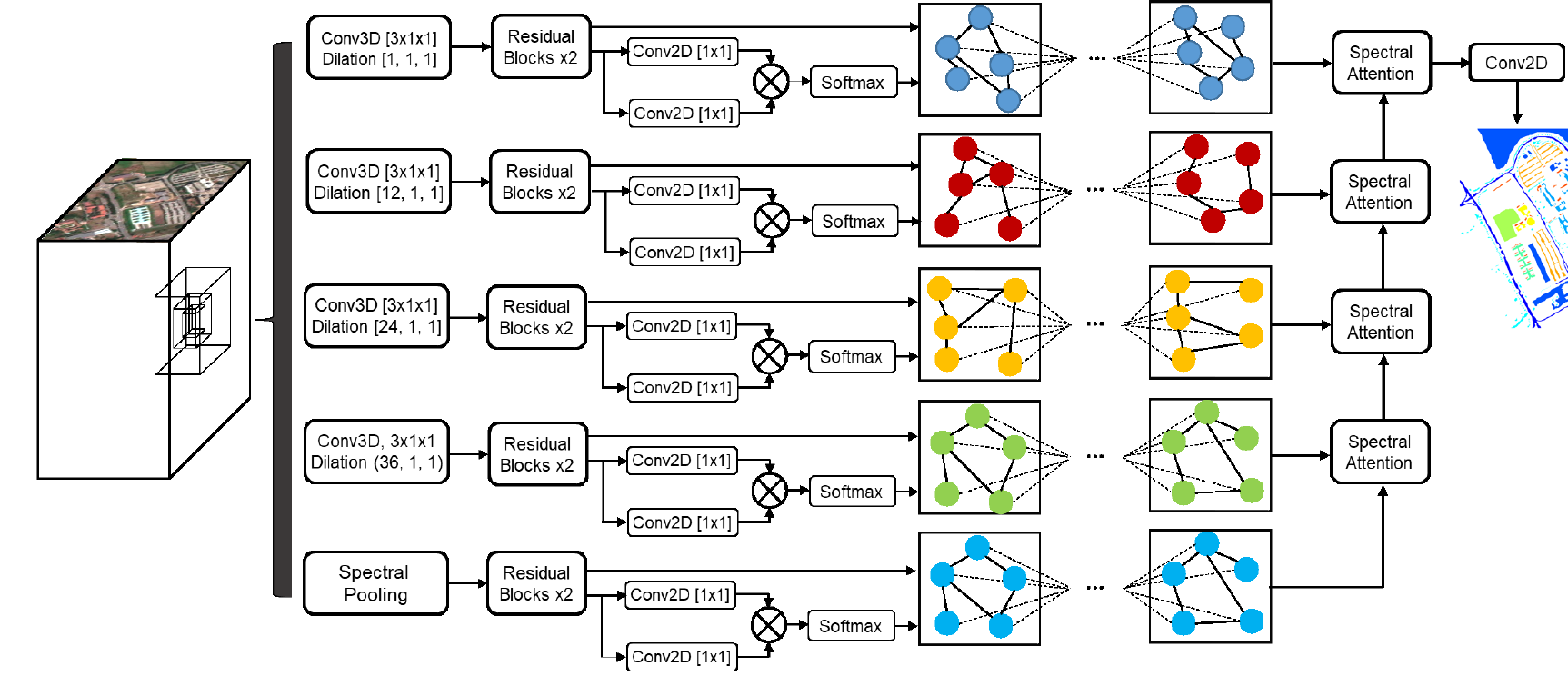}}
\caption{Illustration of the our spectral pyramid graph attention network (SPGAT).}
\label{diagram}
\end{figure*}

\section{Introduction}
The rapid development of hyperspectral sensors enables the observation of  hundreds of continuous bands throughout the electromagnetic spectrum with high spectral resolution. The rich spectral signatures of hyperspectral images (HSIs) facilitate the study of the chemical properties of scene materials remotely. Hyperspectral image classification has consequently been playing an increasingly important role in various fields, such as mining, agriculture, environmental monitoring, and land-cover mapping.

Various approaches have been proposed to address the hyperspectral image classification problem. Early approaches mainly adopted traditional machine learning methods, \emph{e.g.,} KNN \cite{ma2010local}, SVM \cite{kuo2010spatial}, graphical model \cite{shi2012supervised}, extreme learning machine \cite{li2015local}, dictionary learning \cite{chen2011hyperspectral}, and among others trained on hand-crafted features from HSI data to fully exploit the spectral information. Nonetheless, accurately classifying different land-cover categories using only the spectral information regardless of the spatial constraint is difficult. Methods \cite{zhong2014discriminant,zhang2016simultaneous} exploiting both spatial and spectral information have also been proposed to address this issue. 

Early methods largely relied on hand-crafted features empirically designed for HSI data which have limited discriminative power. Inspired by the success of deep neural networks (DNN) in natural image classification, deep learning based methods have been proposed for hyperspectral image classification which have significantly boosted the performance thanks to the strong representation capability. The first DNN approach was proposed by Chen \emph{et al.} \cite{chen2014deep}, which utilized stacked autoencoders to learn high-level features. Recurrent Neural Network (RNN) based architecture was proposed by Mou \emph{et al.} \cite{mou2017deep}. More powerful end-to-end Convolutional Neural Network (CNN) \cite{chen2016deep,hao2017two,hamida20183,wu2019multiscale,chu2019convolution,wang2019hyperspectral,djerriri2019improving,hu2019spnet} based architectures have advanced the state-of-the-art recently. Lee \emph{et al.} \cite{lee2017going} explored local contextual interactions by jointly exploiting local spatio-spectral relationships of neighboring individual pixel vectors. Song \emph{et al.} \cite{song2018hyperspectral} introduced residual learning \cite{he2016deep} to build very deep network for extracting more discriminative features for HSI classification. 

Despite of the remarkable performance by the CNN-based methods, they suffer from several drawbacks. Specifically, standard convolutional kernels work in regular receptive fields for feature response, whose weights are fixed given their position within the small convolution window. Such position-determined weights lead to the isotropy of the kernel with respect to the feature attributes of neightboring locations. This limitation of the standard convolutional kernel neglects the intrinsic or extrinsic structural connections between data points, resulting in poor region delineation and small spurious predictions. To address this problem, graph convolutional network (GCN) based approach $\text{S}^2$GCN has been proposed by Qin \emph{et al.} \cite{QinSTWZT19} which operated on a graph constructed on the local data points and is able to aggregate and transform feature information from the neighbors of every graph node given their relative spatial distance. However, there are several limitations of this approach. Firstly, the graph is constructed based on the rigid pixel lattice which inherently limited the scope of interaction between data points. Secondly, the relation between nodes only accounts for their spatial distance neglecting their correlation in the feature space. Furthermore, $\text{S}^2$GCN only conducts reasoning at a single feature space which is potentially limited to the capacity of the early feature extraction block of the network. Thirdly, $\text{S}^2$GCN is a graph-based semi-supervised approach which predicts unlabeled data with the presence of labeled data which potentially harms its accuracy on dataset without any available labels. 

To explicitly address the above issues, we presents a novel architecture to encode the multiple spectral contextual information in a hierarchical manner, forming multiple spectral embedding spaces. In multiple spectral embedding spaces, we propose graph attention mechanism to explicitly perform context based reasoning in the spatial domain based on the connection in spectral feature space. Our graphs are dynamically constructed based on the intrinsic structure of the data, \emph{i.e.,} nodes of a same neighborhood are assigned with different importance, enabling thorough interaction between comprising data points while increasing the model capacity. After the graph reasoning, an attention based aggregation reminiscent of \emph{multi-head attention} \cite{vaswani2017attention} with focus on specific spectral channels given the contextual scopes is applied on the spectral feature pyramid rather than a simply \textit{averaging} in GAT \cite{VelickovicCCRLB18}.  This is based on the observation that features from different contextual level carry different degrees of discriminative power for classification. Extracting the key discriminative features from each contextual level and enforcing this information during the aggregation can effectively improve the classification accuracy.

\section{Method}

We propose a novel end-to-end graph attention architecture for spectral feature learning and reasoning of HSIs, which consists of mainly two blocks,  \emph{i.e.,} spectral feature pyramid learning and graph attention based reasoning, as illustrated in Fig. \ref{diagram}.

\subsection{Spectral Pyramid Learning}

We propose to utilize $\textit{atrous}$ convolution to probe the HSI signal along the spectral dimension at multiple sampling rates for capturing varying spectral contextual information, forming a spectral pyramid.
It brings us mainly two advantages: (1) multiple spectral contextual information is captured without introducing large convolutional kernel or fully connected layer which significantly 
reduces the number of trainable model parameters and computational complexity (2) large effective fields-of-views is achieved without resorting to resampling features which in turn causes downsampled feature map.

\begin{figure}[htbp]
\centerline{\includegraphics{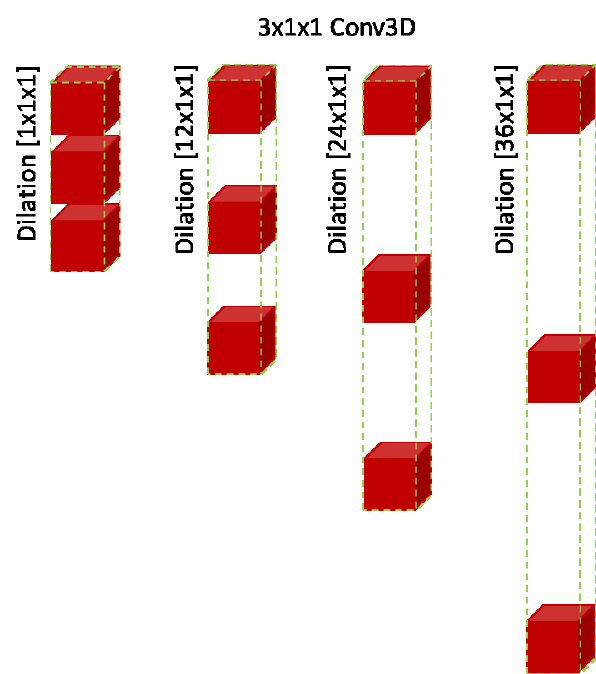}}
\caption{Illustration of $\textit{atrous}$ convolutional kernels along the spectral dimension.}
\label{aspp}
\end{figure}

As illustrated in Fig. \ref{aspp}, we adopt 3D convolutions of kernel size $3 \times 1 \times 1$ and dilation rates $\{(1,1,1), (12,1,1), (24,1,1), (36,1,1)\}$ respectively. 
With the spatial kernel size $1 \times 1$, the 3D convolution is equivalent to 1D convolution over the spectral vector at each pixel location. The output $y[s,i,j]$ of the $\textit{atrous}$ convolution
at signal location $(s,i,j)$ with a filter $w[k]$ of length $K$ is defined as 
\begin{align}
y(s,i,j) = \sum_{k=1}^K  x[s + r\cdot k, i, j] w[k]
\label{eq:1}
\end{align}
where $r$ indicates the stride with which we sample the input signal, and $r=1$ corresponds to the standard convolution.  

A spectral pooling layer is also proposed to gather the global spectral context. Specifically, the spectral pooling layer consists of $[\text{AdaptiveAvgPool3d}+\text{Conv3d}[1\times1\times1]+\text{BatchNorm3d}+\text{ReLU}]$, inspired by \cite{chen2018encoder} in the spatial domain. As illutrated in Fig. \ref{diagram}, the $\textit{atrous}$ spectral pyramid comprises 5 streams, \emph{i.e.,} feature maps from 4 Conv3D with different dilation rates and spectral level features from spectral pooling. The spectral pyramid thus captures spectral contextual information at multiple scales while maintaining a very low computational complexity. 

Residual blocks \cite{he2016deep} are applied on each feature stream to further aggregate local spatial and spectral contexts and transform feature embeddings. Specifically, two bottleneck modules with filter size (64, 128) respectively and expansion factor of 4 are applied per feature stream. 

 $\mathbf{h}'=\{h_{1}'\}\in \mathbb{R}^{d'}$,

\subsection{Graph Attention based Reasoning}
Consider a graph $\mathcal{G(V,E)}$ constructed from a set of node features $\mathbf{h}=\{h_1, h_2,\dots,h_N\}\in \mathbb{R}^d$, where $N$ is the number of nodes and $d$ is the number of features in each node. 
The graph attention layer generates as its output a new set of node features $\mathbf{h}'=\{h_{1}', h_{2}',\dots,h_{N}'\}\in \mathbb{R}^{d'}$, $h_{i}' \in \mathbb{R}^{d'}$. Two 1-D convolutionary 
layers, $\theta(\cdot)$ and $\phi(\cdot)$ are applied on the input feature map respectively in order to initially transform the input features into two sets of higher-level features while obtaining sufficient expressive power.  
Unlike the relatively fixed neighboring relation in pixel lattice, the graph attention layer should be able to dynamically adjust to varying graph structures. 

Specifically, a data-dependent graph which learn a unique graph for each input feature map is formed by determining the connection as well as its strength between two nodes,
\begin{align}
\alpha_{i,j} = a(\theta(h_i), \phi(h_j))
\label{eq:1}
\end{align}
where $a(\cdot)$ is an attention mechanism to compute the attention weight of node $j$ to node $i$. 

Here, we utilize the dot product followed by a linear transformation $\psi$, a LeakyReLU nonlinearity and a $softmax$ operation to measure the normalized attention between two nodes in an embedding space, 
\begin{align}
\alpha_{i,j} = \frac{e^{ \text{LeakyReLU} (\mathbf{W_{\psi}} h_i^T \mathbf{W_{\theta}^T} \mathbf{W_{\phi}} h_j }) }{\sum_{j=1}^{N} e^{ \text{LeakyReLU} (\mathbf{W_{\psi}} h_i^T \mathbf{W_{\theta}^T} \mathbf{W_{\phi}} h_j }) },
\label{eq:2}
\end{align}
where $\mathbf{W_{\theta}}$, $\mathbf{W_{\phi}}$ and $\mathbf{W_{\psi}}$ are the trainable parameters of embedding functions $\theta(\cdot)$, $\phi(\cdot)$ and $\psi$ respectively. To make the attention weights comparable across nodes, they are normalized by $softmax$ operation. As opposed to GAT \cite{VelickovicCCRLB18}, which computes the attention weights based on the concatenation of a pair of features, we compute the feature differences which is more efficient and explicit to characterize the correlation between features. 

Finally, the normalized attention weights are used to perform a weighted combination all nodes to obtain a new set of features,
\begin{align}
h^{'}_{i} =  \text{LeakyReLU} (\sum_{j=1}^{N} \alpha_{i,j} \mathbf{W_{\xi}} h_j),
\label{eq:3}
\end{align}
where $\mathbf{W_{\xi}}$ indicates the weights of a 1-D convolutionary layer $\xi(\cdot)$ before applying a LeakyReLU nonlinearity. 

\begin{figure*}[htbp]
\centerline{\includegraphics{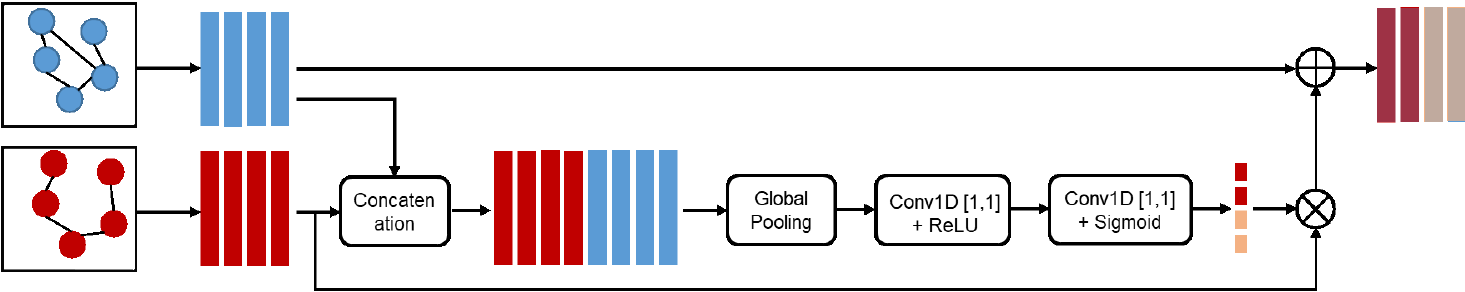}}
\caption{Illustration of spectral attention module.}
\label{ca}
\end{figure*}

After obtaining a new set of feature streams from the multiple graph attention layers, we propose an attention based aggregation with focus on specific spectral channels given the contextual scopes, rather than a simply \textit{averaging} in GAT \cite{VelickovicCCRLB18}. As illustrated in Fig. \ref{ca}, spectral attention module takes the feature maps from
neighboring contextual levels as input and computes spectral level attention coefficients to guide the upper level feature to focus on certain spectral channels. This is based on the observation that features from different contextual level carry different degrees of discriminative power for classification. Extracting the key discriminative features from each contextual level and enforcing this information during the aggregation can effectively encode the contextual information at multiple scales.

\section{Experimental Results}
In this section, we conduct extensive experiments on three publicly available hyperspectral image datasets, and present results using four metrics including per-class accuracy, overall accuracy (OA), average accuracy (AA), and Kappa coefficient. The network architecture of our proposed SPGAT is identical for all the datasets. Specifically, two graph attention layers per contextual level are used, with graphs constructed on 7$\times$7 image patches. Learning rate of 0.001 and Adam optimizer are adopted. We follow \cite{zhou2019} for selecting the training and testing sets.  The number of training epochs is set to 500. All the reported accuracies are calculated based on the average of ten training sessions to obtain stable results. 

\subsection{Datasets}

The University of Pavia dataset captured the University of Pavia with the ROSIS sensor in 2001. It consists of 610$\times$340 pixels with a spatial resolution of 1.3 m$\times$1.3 m and has 103 spectral channels in the wavelength range from 0.43 $\mu$m to 0.86 $\mu$m after removing noisy bands. This dataset includes 9 land-cover classes as listed in Table \ref{table:pu}. The false color image and ground-truth map are shown in Fig. \ref{fig:pu-result}.

\begin{figure*}[!ht]
	\centering
	%\begin{center}
	%\captionsetup[subfigure]{justification=centering}
	\begin{subfigure}{0.23\textwidth}
		\includegraphics[height=5cm,width=0.99\linewidth]{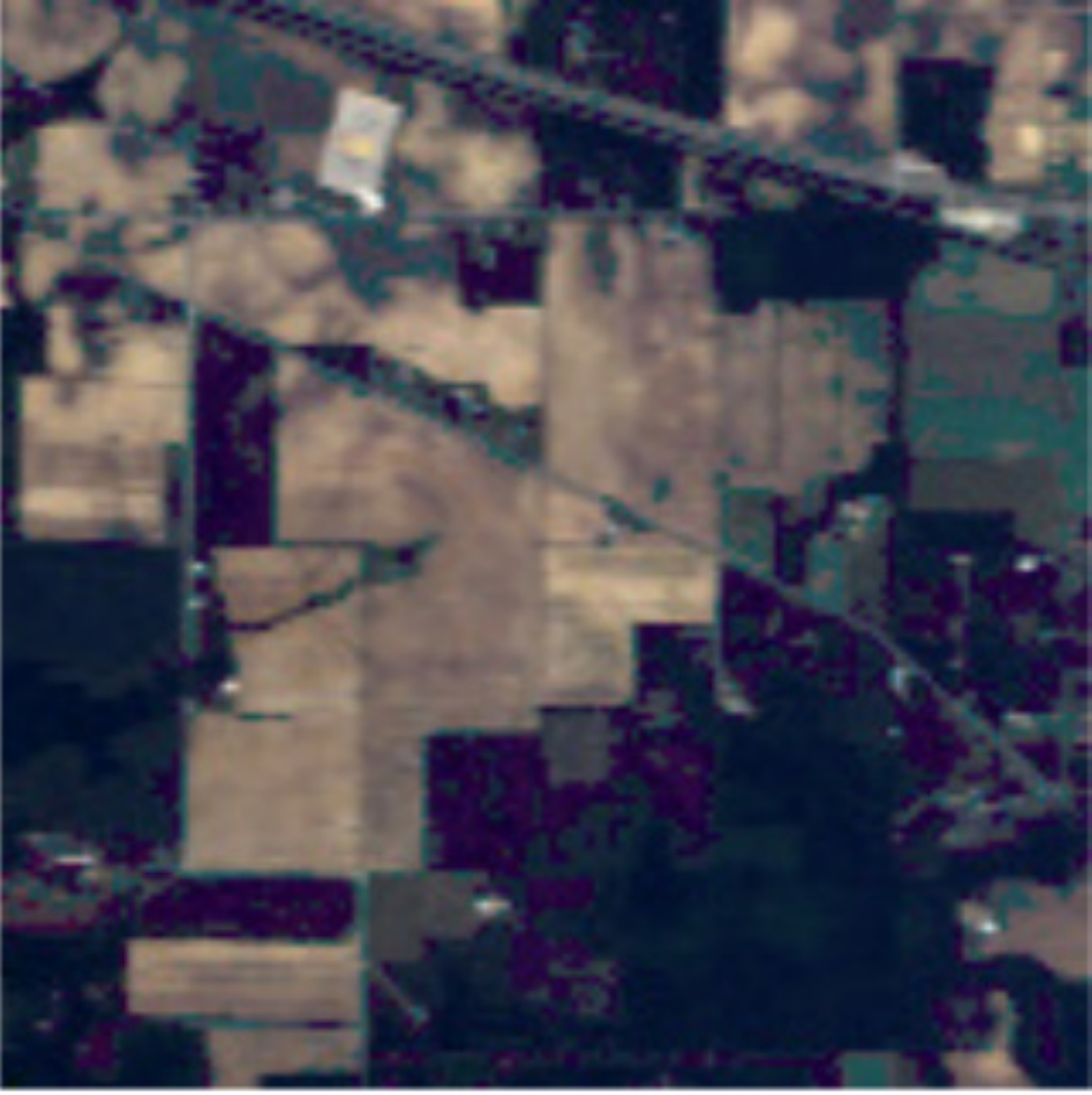}
		%\caption{(a)}
		\label{fig:Image}
	\end{subfigure}%
	%\hfill %%useful if width of each figure is less the 
	\begin{subfigure}{0.23\textwidth}
		\includegraphics[height=5cm,width=0.99\linewidth]{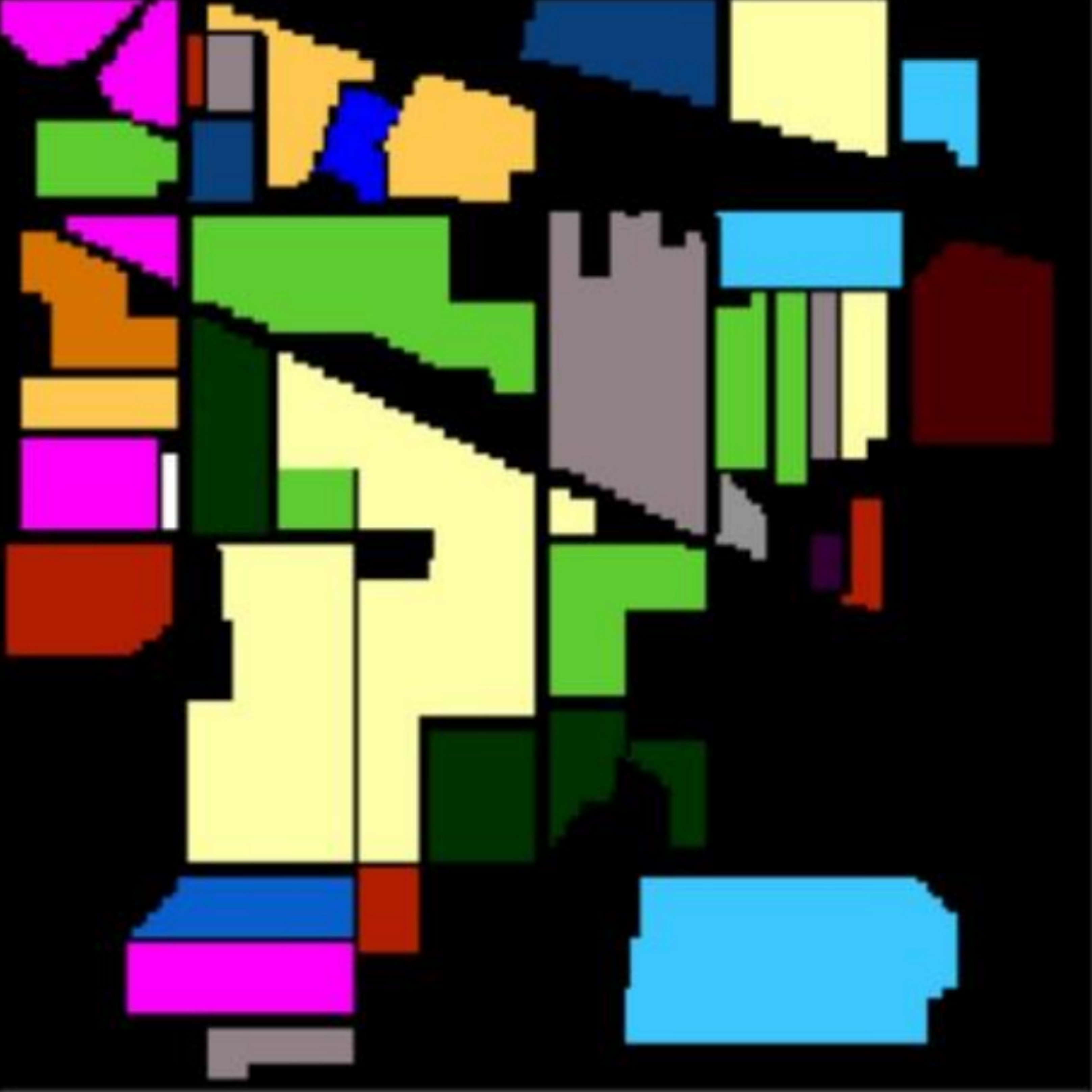}
		%\caption{(b)}
		\label{fig:Deeplab_v3_plus_16}
	\end{subfigure}%
	%\hfill % <-- added
	\begin{subfigure}{0.23\textwidth}
		\includegraphics[height=5cm,width=0.99\linewidth]{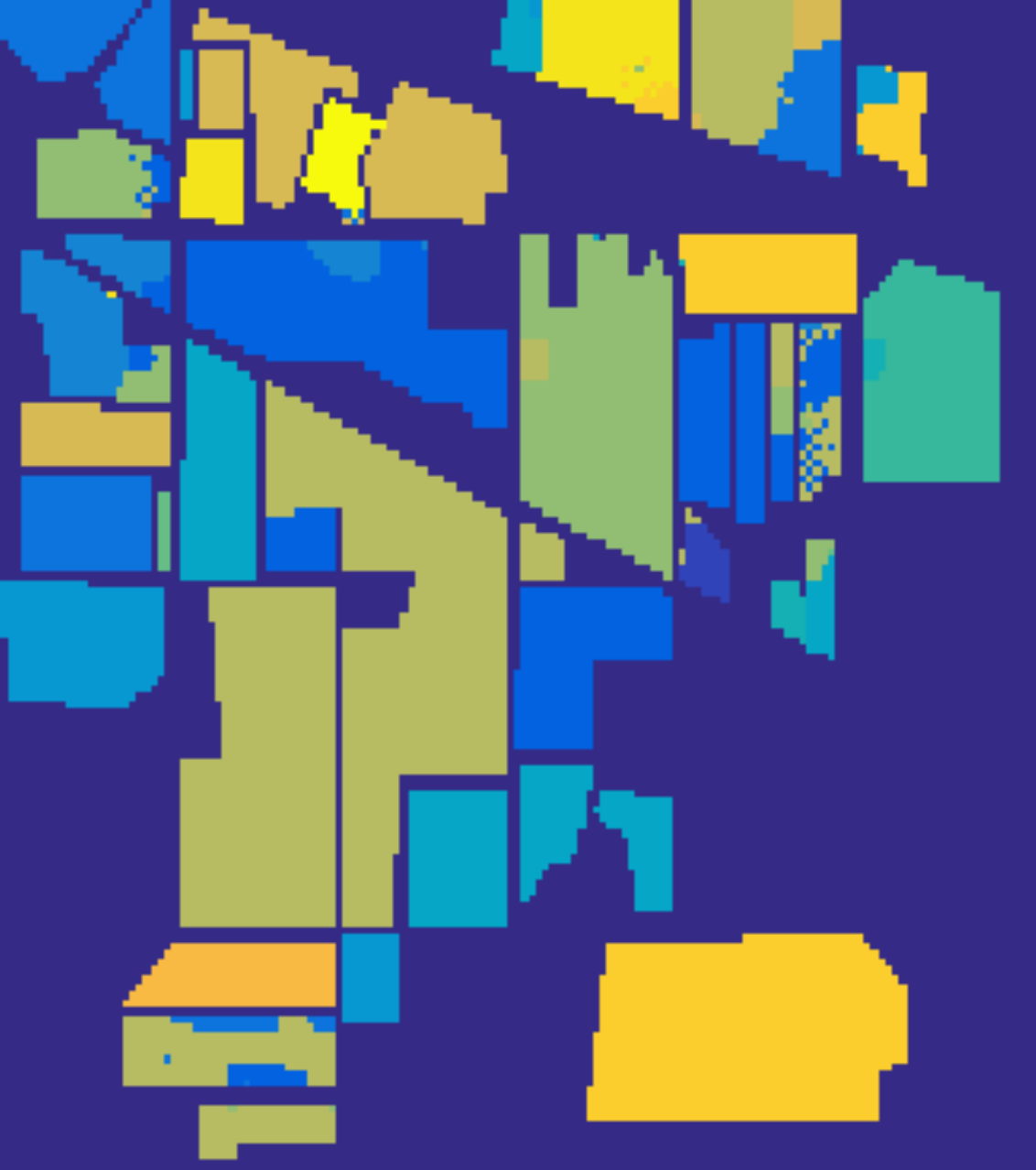}
		%\caption{(c)}
		\label{fig:MeshNet}
	\end{subfigure}
	\begin{subfigure}{0.23\textwidth}
		\includegraphics[height=5cm,width=0.99\linewidth]{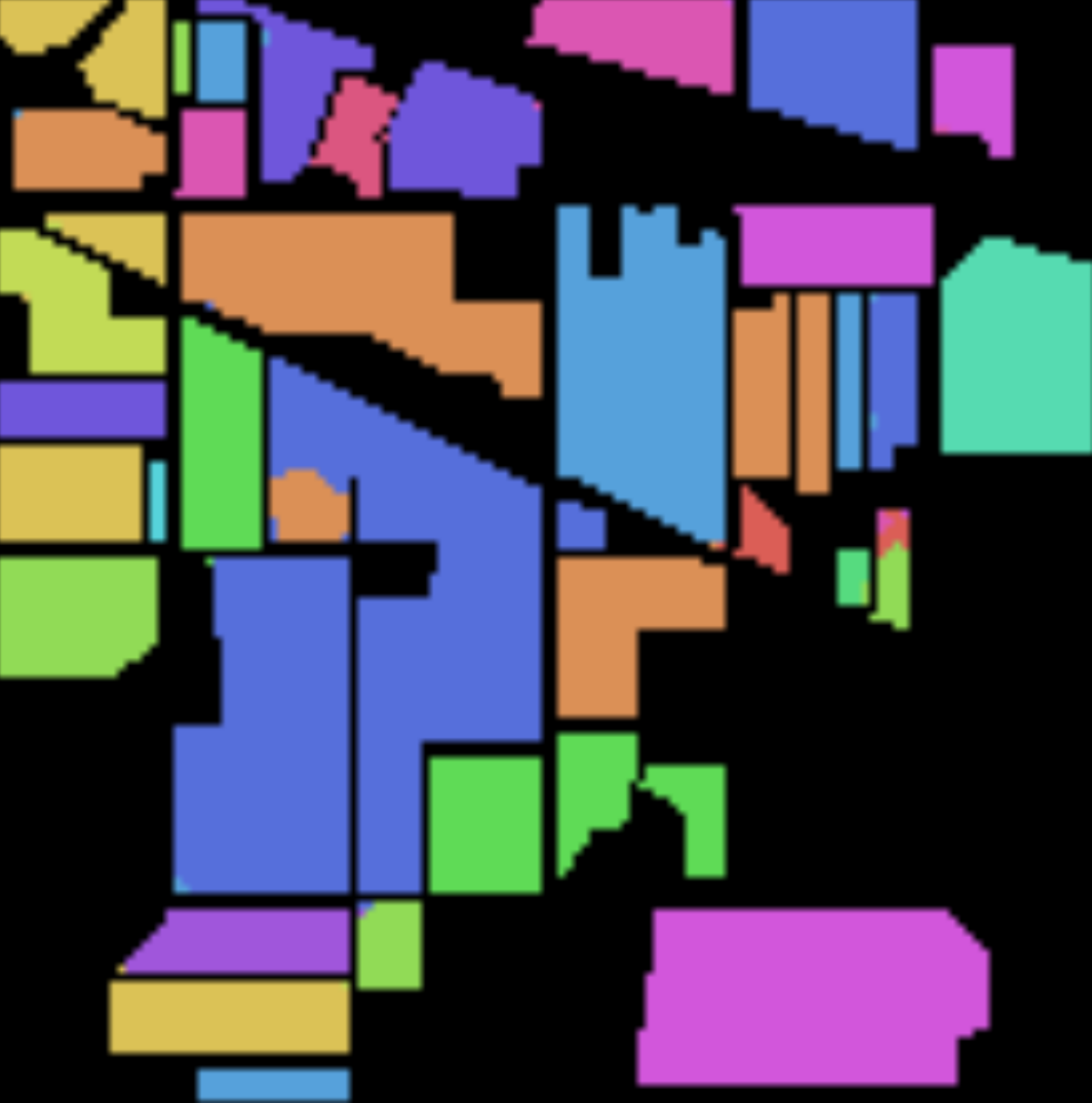}
		%\caption{(c)}
		\label{fig:MeshNet}
	\end{subfigure}
	\caption{Indian Pines dataset: (a) False color image (b) Groundtruth map (c) Prediction of LBMSELM \cite{CaoYRCHS19} (d) Prediction of the proposed SPGAT}.
	\label{fig:indianpines-result}
\end{figure*}

The Indian Pines dataset was collected by Airborne Visible/Infrared Imaging Spectrometer sensor which consists of 145$\times$145 pixels with a spatial resolution of 20 m$\times$20 m and has 220 spectral channels covering the range from 0.4 $\mu$m to 2.5 $\mu$m. Fig. \ref{fig:indianpines-result} exhibits the false color image and ground-truth map of the Indian Pines dataset. This dataset includes 16 land-cover classes as listed in Table \ref{table:ip}.

\begin{figure*}[!ht]
	\centering
	%\begin{center}
	%\captionsetup[subfigure]{justification=centering}
	\begin{subfigure}{0.23\textwidth}
		\includegraphics[height=5cm,width=0.99\linewidth]{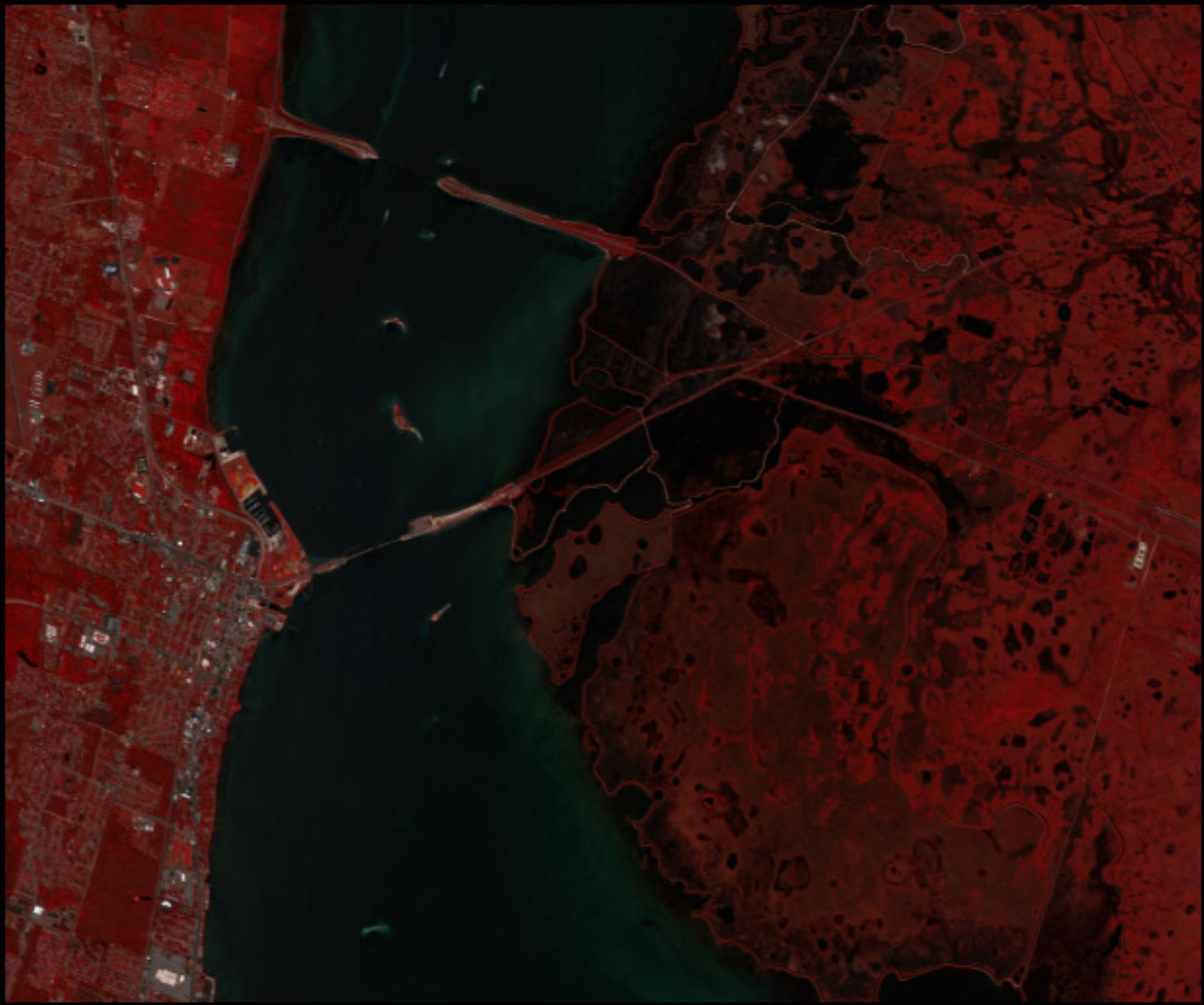}
		%\caption{(a)}
		\label{fig:Image}
	\end{subfigure}%
	%\hfill %%useful if width of each figure is less the 
	\begin{subfigure}{0.23\textwidth}
		\includegraphics[height=5cm,width=0.99\linewidth]{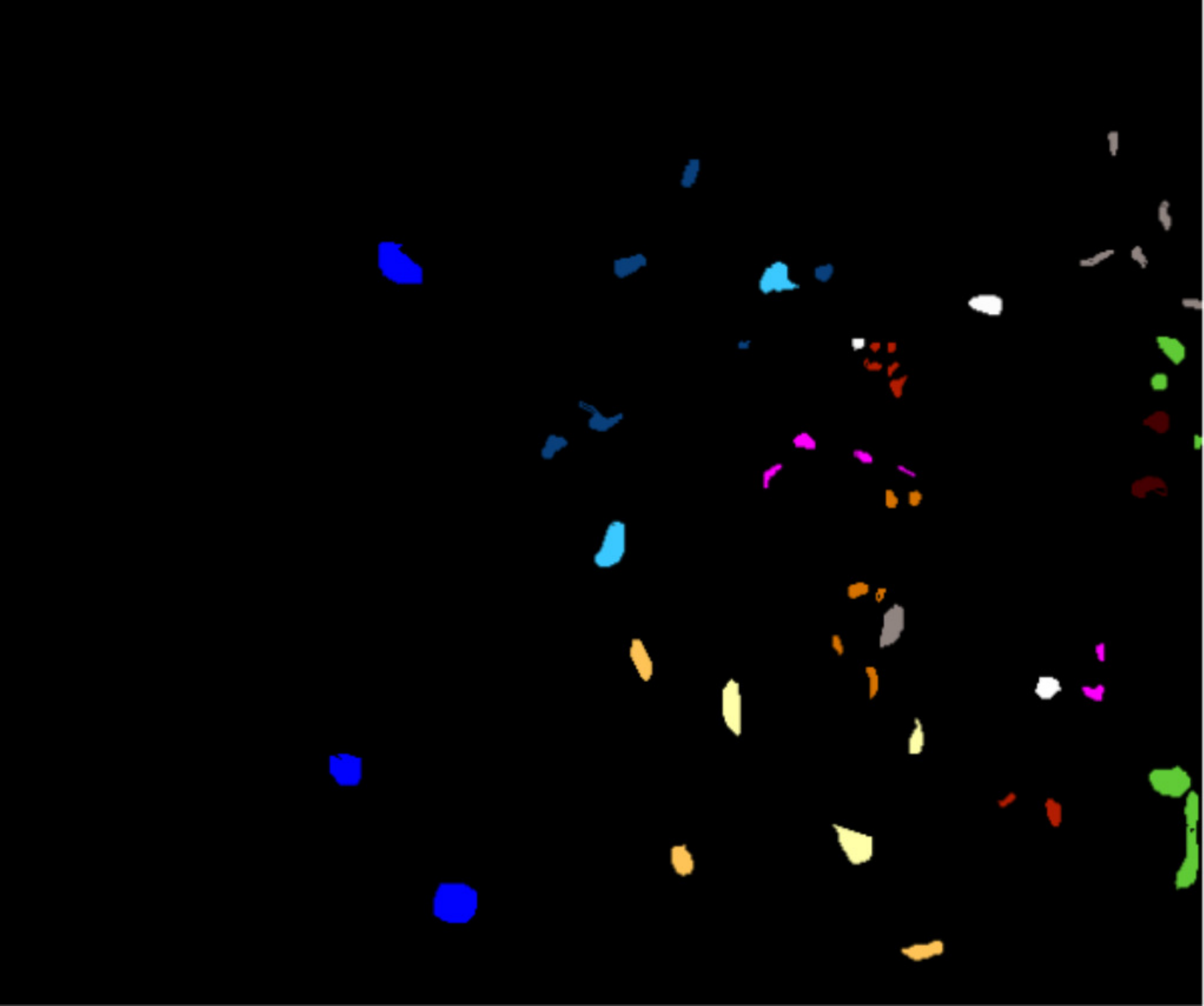}
		%\caption{(b)}
		\label{fig:Deeplab_v3_plus_16}
	\end{subfigure}%
	%\hfill % <-- added
	\begin{subfigure}{0.23\textwidth}
		\includegraphics[height=5cm,width=0.99\linewidth]{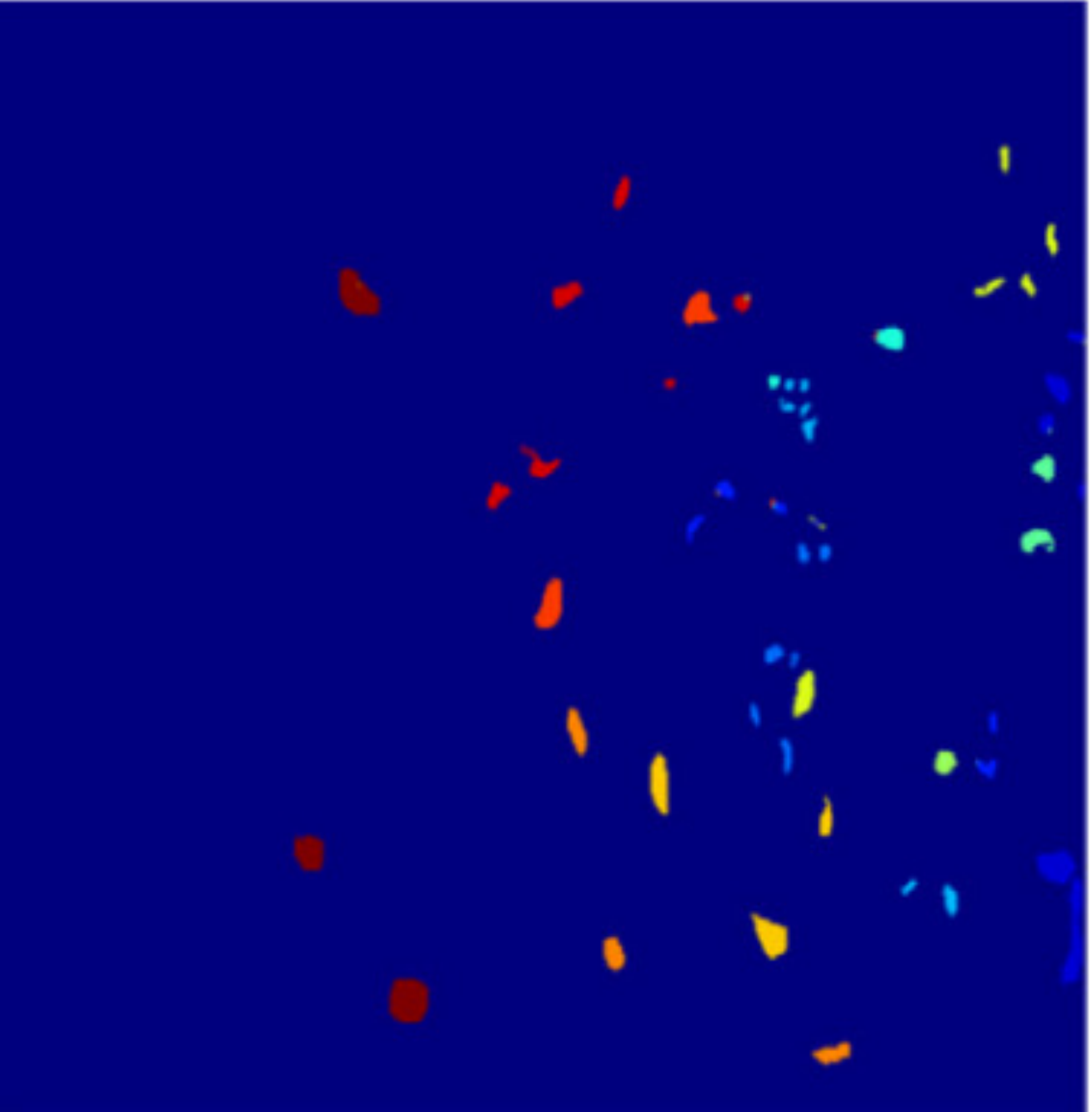}
		%\caption{(c)}
		\label{fig:MeshNet}
	\end{subfigure}
	\begin{subfigure}{0.23\textwidth}
		\includegraphics[height=5cm,width=0.99\linewidth]{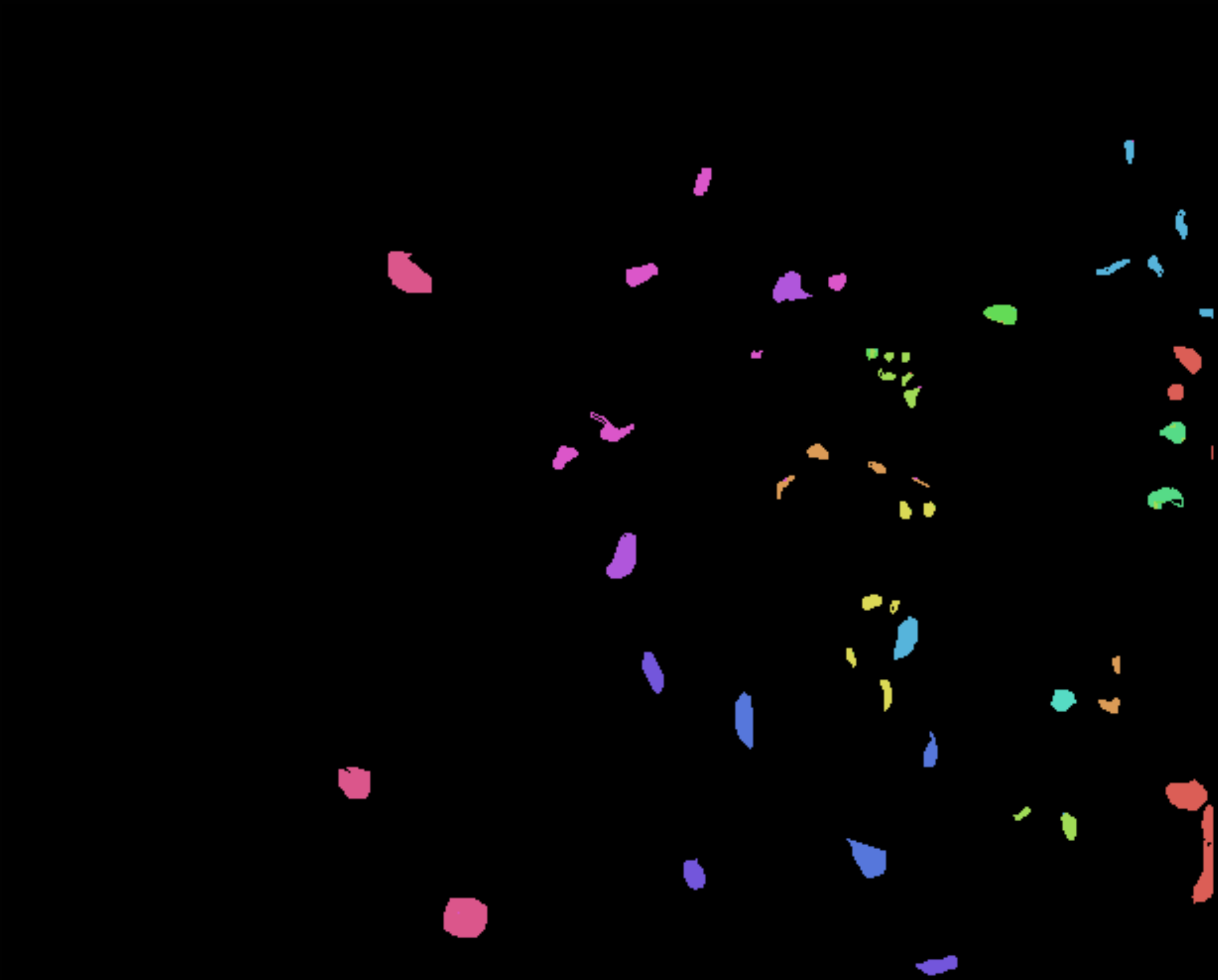}
		%\caption{(c)}
		\label{fig:MeshNet}
	\end{subfigure}
	\caption{Kennedy Space Center dataset: (a) False color image (b) Groundtruth map (c) Prediction of SSLSTMs \cite{zhou2019} (d) Prediction of the proposed SPGAT}
	\label{fig:ksc-result}
\end{figure*}

The Kennedy Space Center dataset which was taken by AVIRIS sensor over Florida with a spectral coverage ranging from 0.4 $\mu$m to 2.5 $\mu$m, contains 224 bands and 614 $\times$ 512 pixels with a spatial resolution of 18 m. This dataset comprises 13 land-cover classes as listed in Table \ref{table:ksc}, and Fig. \ref{fig:ksc-result} exhibits the false color image and ground-truth map.

\subsection{Classification Results}

We quantitatively and qualitatively evaluate our proposed method and compare with various recent deep learning based methods to demonstrate its effectiveness.
\subsubsection{The University of Pavia Dataset}
Table \ref{table:pu} presents the quantitative results obtained by different methods on the University of Pavia dataset, where the highest value in each row is highlighted in bold. 
Our proposed SPGAT outperforms the competing methods on 7 out of 9 categories and exhibits the best overall results, \emph{i.e.,} OA, AA and Kappa. In general, all methods except SSLSTMs \cite{zhou2019} and SPGAT 
fail to capture larger context, which consequentially produce inferior results. SSLSTMs applies LSTM to learn both spatial and spectral features, however it fails to characterize the 
large scale spectral contexts due to the limitation of LSTM models on high dimension feature. On the contrary, our SPGAT encodes the spectral contexts in a hierarchical manner and utilizes
graph attention models to explicitly perform multi-scale inference in the spatial domain based on the connection in spectral feature space. 

\begin{figure*}[!ht]
	\centering
	%\begin{center}
	%\captionsetup[subfigure]{justification=centering}
	\begin{subfigure}{0.23\textwidth}
		\includegraphics[height=5cm,width=0.99\linewidth]{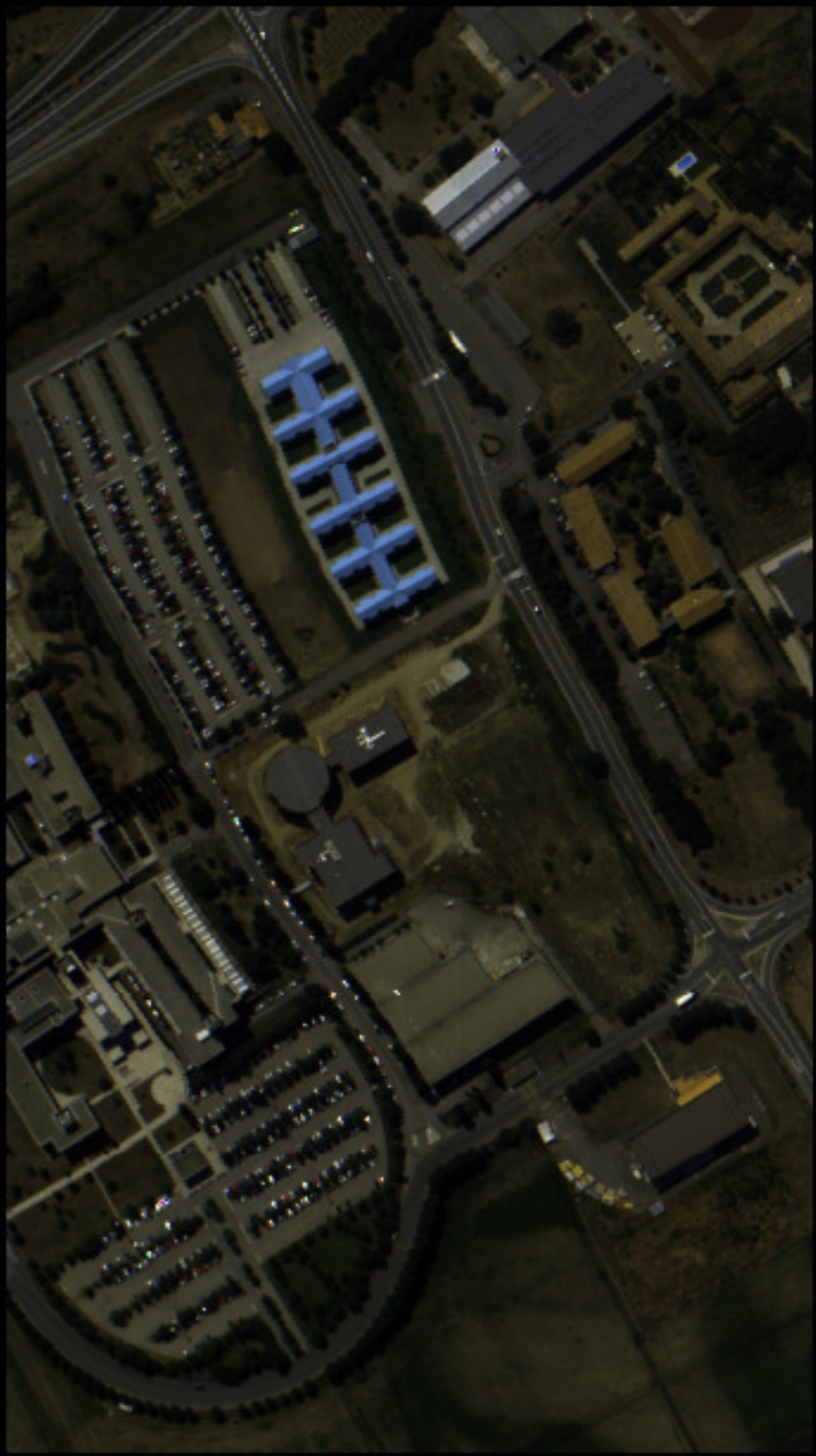}
		%\caption{(a)}
		\label{fig:Image}
	\end{subfigure}%
	%\hfill %%useful if width of each figure is less the 
	\begin{subfigure}{0.23\textwidth}
		\includegraphics[height=5cm,width=0.99\linewidth]{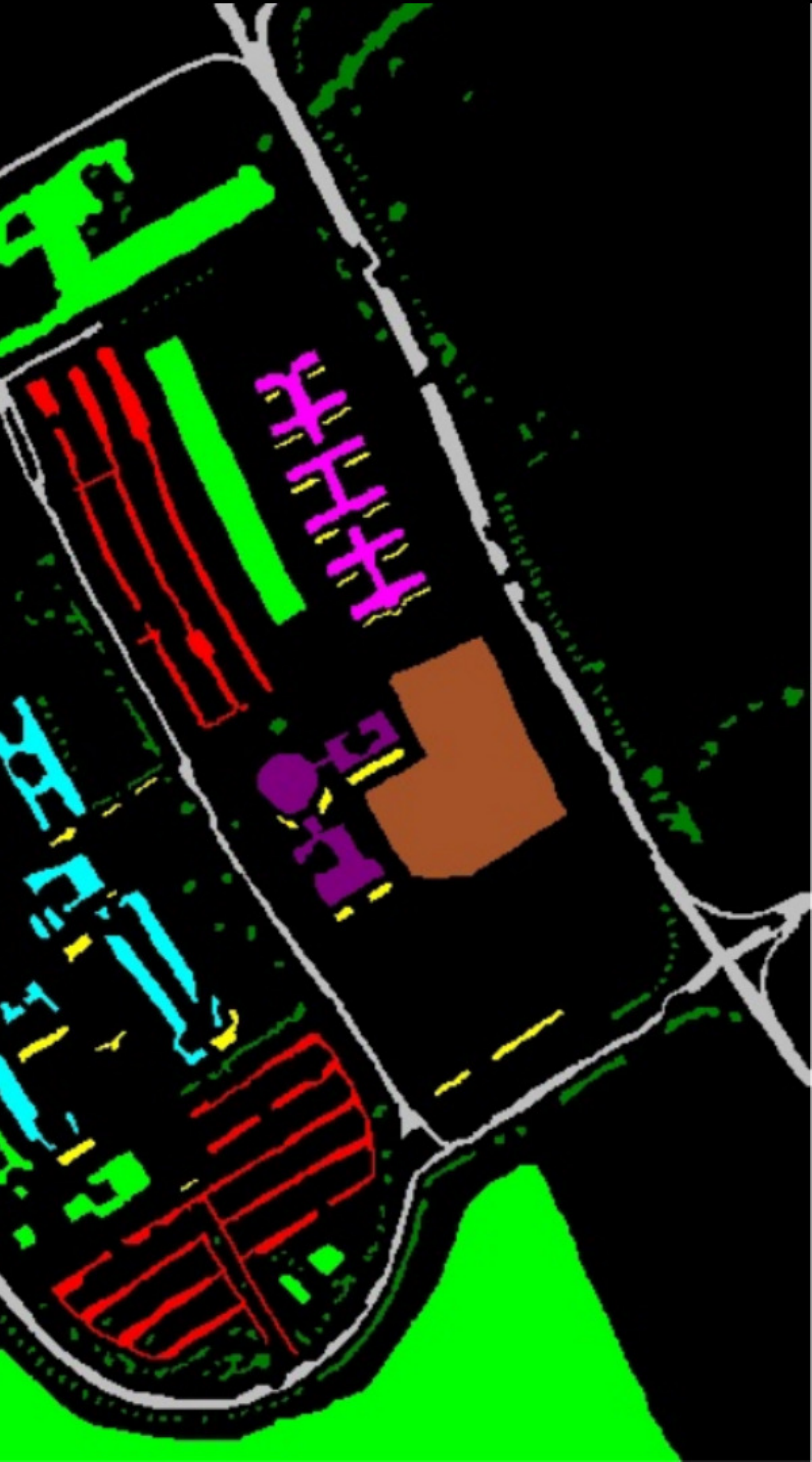}
		%\caption{(b)}
		\label{fig:Deeplab_v3_plus_16}
	\end{subfigure}%
	%\hfill % <-- added
	\begin{subfigure}{0.23\textwidth}
		\includegraphics[height=5cm,width=0.99\linewidth]{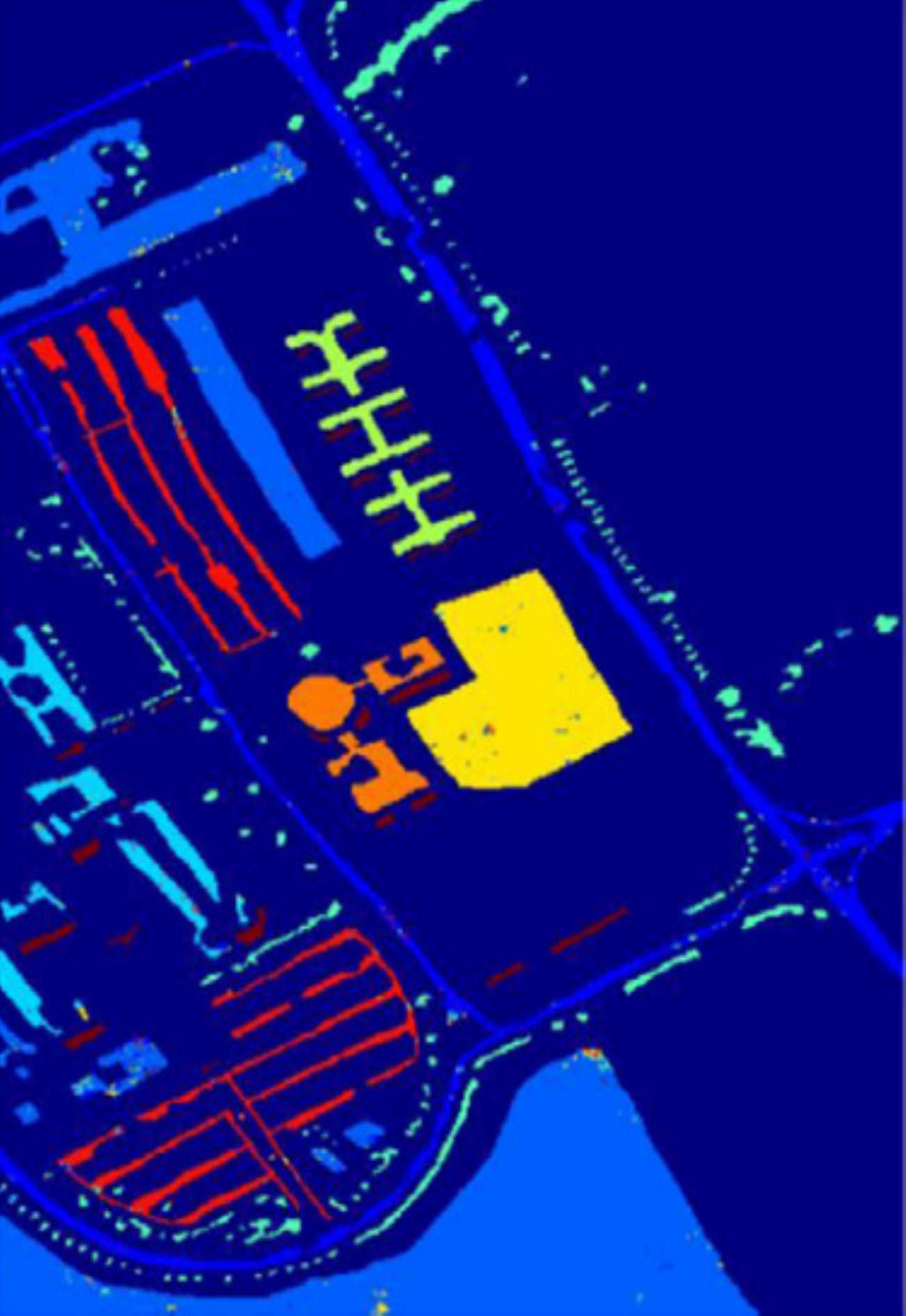}
		%\caption{(c)}
		\label{fig:MeshNet}
	\end{subfigure}
	\begin{subfigure}{0.23\textwidth}
		\includegraphics[height=5cm,width=0.99\linewidth]{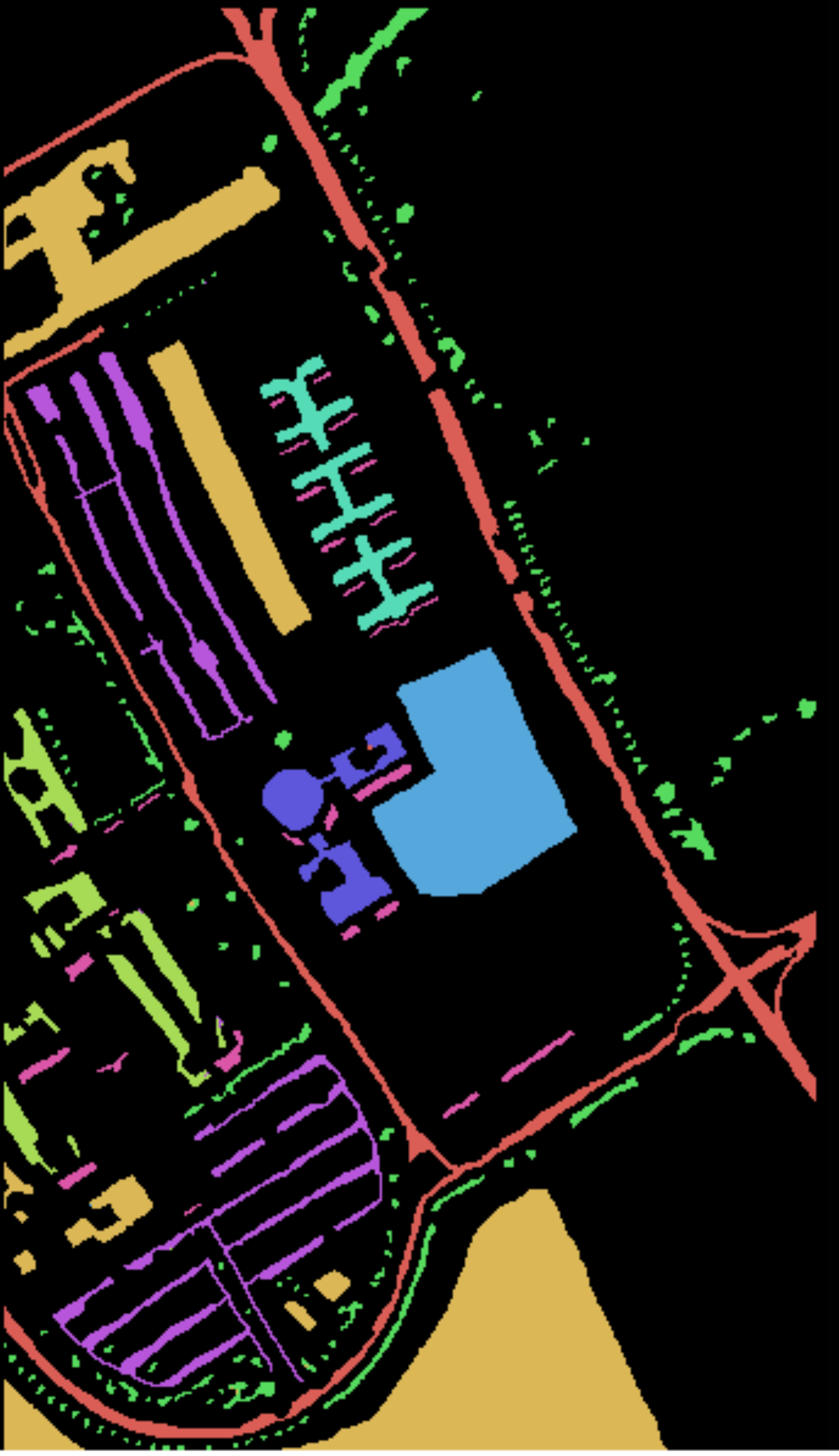}
		%\caption{(c)}
		\label{fig:MeshNet}
	\end{subfigure}
	\caption{The university of Pavia dataset: (a) False color image (b) Groundtruth map (c) Prediction of SSLSTMs \cite{zhou2019} (d) Prediction of the proposed SPGAT}
	\label{fig:pu-result}
\end{figure*}

Fig. \ref{fig:pu-result} shows a visual comparison of SPGAT and SSLSTMs on the University of Pavia dataset. We can observe that SSLSTMs suffers from mis-classifications even in large regions of the same category, due to the lack of 
inference in the spatial space as well as the failure to encode the most discriminative spectral features. We can see that our SPGAT is able to produce both accurate and coherent predictions.

\subsubsection{The Indian Pines Dataset}
Table \ref{table:ip} presents the quantitative results obtained by different methods on the Indian Pines dataset. In addition to tradition methods, we also compare with methods
based on graph neural networks, \emph{i.e.,} GCN \cite{KipfW17} and $\text{S}^2$GCN \cite{QinSTWZT19}, to demonstrate the effectiveness of our method. Overall, our method surpasses 
all the compared methods, with a significant margin of $5.11\%$ comparing to the best competing method $\text{S}^2$GCN. GCN and $\text{S}^2$GCN both construct their graphs based the pixel lattice, whereas
$\text{S}^2$GCN encodes the spatial distance between adjacent nodes in graph which enables it to caputure the local instrinsic structure in HSI data. However, SPGAT has the advantages over the compared GNN based
methods as follows: (a) SPGAT constructs data-driven graph rather than a rigid spatial structure (b) SPGAT assigns adjacency matrix with both connectivities and strengths between nodes which fully characterize
their correlation in spectral feature space (c) SPGAT features the multiple spectral contextual graph based inference which enables robustness and accuracy in generalizing to HSI data captured 
by various sensors.

\begin{table*}
	\centering
	%\begin{center}
	\begin{tabular}{lccccccc}
		\hline
		Class  & R-PCA CNN \cite{makantasis2015} & PPF-CNN \cite{li2016} & CD-CNN \cite{lee2017} & SS-CNN \cite{mei2017learning} & 3D-CNN \cite{chen2016deep}  & SSLSTMs \cite{zhou2019} & SPGAT \\
		\hline
        Asphalt & 92.43 & 97.42 & 94.60 & 97.40 & 96.72 & 96.83 & \textbf{98.60} \\
        Meadows & 94.84 & 95.76 & 96.00 & \textbf{99.40} & 96.31 & 98.74 & 96.11 \\
        Gravel & 90.89 & 94.05 & 95.50 & \textbf{98.84} & 97.15 & 96.57 & 97.83 \\
        Trees & 93.99 & 97.52 & 95.90 & 99.16 & 96.16 & 98.43 & \textbf{99.23} \\
        Painted metal sheets & \textbf{100.0} & \textbf{100.0} & \textbf{100.0} & \textbf{100.0} & 99.81 & 99.94 & \textbf{100.0} \\
        Bare Soil &92.86 & 99.13 & 94.10 & 98.70 & 94.87 & 99.43 & \textbf{100.0} \\
        Bitumen & 93.89 & 98.96 & 96.19 & \textbf{100.0} & 97.44  & 99.31 & \textbf{100.0} \\
        Self-Blocking Bricks & 91.18 & 93.62 & 88.80 & 94.57 & 98.23 & 97.98 & \textbf{99.92} \\
        Shadows & 99.33 & 99.60 & 99.5 & 99.87 & 98.04 & 99.39 & \textbf{100.0} \\
		\hline
		OA & 93.87 & 96.48 & 96.73 & 98.41  & 96.55  & 98.48 & \textbf{98.92}\\
		AA & 94.38 & 97.03 & 95.77 & 98.22 & 97.19 & 98.51 & \textbf{99.07}\\
		Kappa & - & - & - & - &  95.30 & 97.56 & \textbf{97.86}\\
		\hline
		%\vfill
	\end{tabular}
	%\end{center}
	\caption{Per-class accuracy, OA, AA (\%), and Kappa coefficient achieved by different methods on the University of Pavia dataset.}%and \textbf{IS}: deep supervision
	\label{table:pu}
\end{table*}

\begin{table*}
	\centering
	%\begin{center}
	
	\begin{tabular}{lccccccc}
		\hline
		Class & LapSVM \cite{yang2013} & $\text{S}^2$SL \cite{dopido2013} & SSGEL \cite{cao2017embedding} & GCN \cite{KipfW17} & $\text{S}^2$GCN \cite{QinSTWZT19} & LBMSELM \cite{CaoYRCHS19} & SPGAT \\
		\hline
        Alfalfa & 91.81 & 84.96& \textbf{100.0} & 20.03 & \textbf{100.0} & 98.03 & 96.22 \\
        Corn-notill& 72.89& 70.45 & 84.72& 60.92 & 92.22 & 89.97 & \textbf{99.27} \\
        Corn-mintill& 63.45& 65.21 & 84.70& 45.99 & 84.97 & 65.73 & \textbf{94.42} \\
        Corn&  86.78&  85.90& 83.02 & 37.29 & 91.11 & 82.82 & \textbf{98.43} \\
        Grass-pasture& 77.40& 87.81& 84.05& 89.08 & \textbf{100.0} & 81.89 & 95.86 \\
        Grass-trees& 96.39& 96.30& 90.28 & 84.83 & 99.18 & 98.20 & \textbf{99.95} \\
        Grass-pasture-mowed& \textbf{100.0} & 91.24&  \textbf{100.0} & \textbf{100.0} & 98.2& 100.0 & 95.78 \\
        Hay-windrowed& 97.25& 98.51 & \textbf{100.0} & 51.76 & 97.53 & 99.69 & \textbf{100.0} \\
        Oats& \textbf{100.0} & 95.55& 26.30 & 61.01 & \textbf{100.0} & \textbf{100.0} & 96.80 \\
        Soybean-notill& 73.44& 83.91& 76.68& 65.90 & 97.41 & 74.49 & \textbf{98.15} \\
        Soybean-mintill& 61.67& 69.23& 86.34& 51.57 & 83.95 & 92.38 & \textbf{98.29} \\
        Soybean-clean& 66.55& 83.47& 66.10& 77.04 & 91.35 & 81.86 & \textbf{98.79} \\
        Wheat& 98.41& 99.05 & 99.03 & 70.91 & 100.0 & \textbf{100.0} & 99.88 \\
        Woods& 89.05& 93.81 & \textbf{99.79} & 66.87 & 99.01 & 93.80 & 99.76 \\
        Buildings-Grass-Trees-Drives& 80.92& 67.94& 76.46& 63.38 & 87.19 & 76.10 & \textbf{88.19}\\
        Stone-Steel-Towers& 99.55& 85.81& \textbf{100.0} & 62.22 & 94.85 & 94.94 & 98.93\\
		\hline
		OA & 75.71 & 79.85 & 86.53 & 66.87 & 91.64 & 87.47 & \textbf{96.75}\\
		AA & 84.72 & 84.61 &  83.41& 63.38 & 94.92 & 89.37 & \textbf{97.42}\\
		Kappa & 72.62 & 77.25 &  84.72 & 62.22 & 90.41 & 85.61 & \textbf{96.30}\\
		\hline
		%\vfill
		%\toprule[1pt]
	\end{tabular}
	%\end{center}
	\caption{Per-class accuracy, OA, AA (\%), and Kappa coefficient achieved by different methods on the Indian Pines dataset}%and \textbf{IS}: deep supervision
	\label{table:ip}
\end{table*}

\subsubsection{The Kennedy Space Center Dataset}
Table \ref{table:ksc} presents the quantitative results obtained by different methods on the Kennedy Space Center dataset. Similar to the results on the University of Pavia dataset,
SPGAT produces the best OA accuracy among all compared methods, and SSLSTMs achieves the second best. It is worth-noting that two other graph based approaches, \emph{i.e.,} GCN \cite{KipfW17}
and ELP-RGF \cite{cui2018semi} also show promising results compared with traditional methods, which demonstrates the advantage of conducting graph reasoning.

\begin{table*}
	\centering
	%\begin{center}
	\begin{tabular}{lccccccc}
		\hline
		Class  & LapSVM \cite{yang2013} & SSLP-SVM \cite{wang2014semi} & GCN \cite{KipfW17}& ELP-RGF \cite{cui2018semi} & SSLSTMs \cite{zhou2019} & SPGAT \\
		\hline
        Scrub & 87.17 & 87.19& 86.91 & \textbf{100.0} & 99.56 & 97.86 \\
        Willow swamp & 95.63 & 77.38 & 83.29 & \textbf{99.75} & 90.41 & 95.86 \\
        Cabbage palm hammock & 70.90 & 85.87& 87.57 &93.06 & \textbf{100.0} & 99.72 \\
        Cabbage palm/oak hammock &83.97 & 51.97& 24.86 &75.49 & \textbf{99.56} & 91.24 \\
        Slash pine &79.08 &41.13 & 63.36 & 55.95& \textbf{93.79} & 89.62 \\
        Oak/broadleaf hammock &  89.62 & 36.43& 61.01 &95.64 & \textbf{95.15} & 92.53 \\
        Hardwood swamp & 96.34 & 72.06& 91.20 &98.79  & \textbf{100.0} & 94.06 \\
        Graminoid marsh & 93.34 &76.47 & 78.20 &99.10 & 88.40 & \textbf{98.65} \\
        Spartina marsh & 98.12 & 89.52& 85.39 &97.21 & 99.57 & \textbf{99.89} \\
        Cattail marsh & 92.90 &75.53  & 84.28 & 84.79& \textbf{100.0} & 99.87 \\
        Salt marsh & 94.92 & 84.47& 94.68 & 99.95& 99.47 & \textbf{99.95} \\
        Mud flats & 94.22 &68.16 & 82.14 & 94.21  & 98.90 & \textbf{99.31} \\
        Water &99.08 &99.15 & 98.99 & \textbf{100.0} & 99.88 & \textbf{100.0} \\
		\hline
		OA & 91.25 &75.52 & 83.60  & 93.21  & 97.89 & \textbf{98.15}\\
		AA & 90.41 &72.72 & 78.60 & 91.84& \textbf{97.28} & 96.81\\
		Kappa & 90.25 & 72.88 & 81.70 &  92.45& 97.65 & \textbf{97.84}\\
		\hline
		%\vfill
		%\toprule[1pt]
	\end{tabular}
	%\end{center}
	\caption{Per-class accuracy, OA, AA (\%), and Kappa coefficient achieved by different methods on the Kennedy Space Center dataset.}%and \textbf{IS}: deep supervision
	\label{table:ksc}
\end{table*}

\subsection{Ablation Study}

In this section, we conduct ablation study on the University of Pavia dataset to investigate the effectiveness of various novel parts of our proposed architecture. To show
the importance of generating multiple spectral contextual feature, we remove the branches where the dilation rate is larger than 1, keeping the branch with dilation rate of 1 (SPGAT-1).
In order to demonstrate the effectiveness of attention graph layer, we compare with the baseline model replacing GAT with GCN \cite{KipfW17} layers (SPGCN). Finally we show that the spectral attention 
block aggregates features preserving the discriminative information across multiple spectral contextual level compared with averaging (SPGAT-Avg). As summarized in Table \ref{table:ablation},
introducing multiple spectral contextual embedding spaces enables an OA gain of $3.67\%$, whilst replacing graph attention layer with GCN leads to OA drop of $2.55\%$. We also observe that
the spectral attention block also boosts the performance by $0.31\%$.

\begin{table*}
	\centering
	%\begin{center}
	\begin{tabular}{lccccccc}
		\hline
		  & SPGAT-1 & SPGCN & SPGAT-Avg & SPGAT \\
		\hline
		OA & 95.25 & 96.37 & 98.61 & 98.92\\
		\hline
%		\vfill
		%\toprule[1pt]
	\end{tabular}
	%\end{center}
	\caption{OA achieved by different baselines on the University of Pavia dataset.}%and \textbf{IS}: deep supervision
	\label{table:ablation}
\end{table*}

\section{Conclusions}
In this paper, we proposed a novel architecture for hyperspectral image classification. We demonstrated that using $\textit{Atrous}$ convolution
to probe the HSI signal along the spectral dimension at multiple sampling rates could efficiently and effectively encode varying spectral contextual 
information. We further proposed graph attention based reasoning in each spectral embedding space which produced significantly boosted classification 
accuracy compared with existing methods.

\bibliographystyle{./IEEEtran}
\bibliography{./conference_arxiv}

\end{document}